\documentclass[preprint,authoryear]{elsarticle}

\usepackage{bbm}
\usepackage{float}
\usepackage{amssymb}
\usepackage{url}
\usepackage[a4paper, margin=0.75in]{geometry} 
\usepackage[linesnumbered,ruled,vlined]{algorithm2e} 
\usepackage{booktabs}
\usepackage{chngcntr}
\usepackage{amsmath}
\counterwithout{table}{section}
\renewcommand{\arraystretch}{1.3}
\usepackage[authoryear]{natbib}         
\usepackage{tcolorbox}

\journal{Transportation Research Part C: Emerging Technologies}

\begin{document}

\begin{frontmatter}

\title{CARE-Drive: A Framework for Evaluating Reason-Responsiveness of Vision–Language Models in Automated Driving}

\author[inst1,inst3]{Lucas Elbert Suryana}
\ead{l.e.suryana@tudelft.nl}
\author[inst4]{Farah Bierenga}
\author[inst4]{Sanne van Buuren}
\author[inst4]{Pepijn Kooij}
\author[inst4]{Elsefien Tulleners}
\author[inst2]{Federico Scari}
\author[inst1,inst3]{Simeon Calvert}
\author[inst1]{Bart van Arem}
\author[inst3,inst2]{Arkady Zgonnikov}

\affiliation[inst1]{organization={Department of Transport \& Planning, Faculty of Civil Engineering and Geosciences, Delft University of Technology},
            }

\affiliation[inst2]{organization={Department of Cognitive Robotics, Faculty of Mechanical Engineering, Delft University of Technology},
            }

\affiliation[inst3]{organization={Centre for Meaningful Human Control, Delft University of Technology},
            }

\affiliation[inst4]{
    organization={Faculty of Mechanical Engineering, Delft University of Technology},
    city={Delft},
    country={The Netherlands}
}

\begin{abstract}
Foundation models, including vision–language models (VLMs), are increasingly used in automated driving to interpret scenes, recommend actions, and generate natural-language explanations. However, existing evaluation methods primarily assess outcome-based performance, such as safety and trajectory accuracy, without determining whether model decisions appropriately reflect human-relevant considerations. As a result, it remains unclear whether explanations produced by such models correspond to genuine reason-responsive decision-making or merely post-hoc rationalizations. This limitation is particularly significant in safety-critical domains because it can create false confidence. To address this gap, we propose CARE-Drive (Context-Aware Reasons Evaluation for Driving), a model-agnostic framework for evaluating reason-responsiveness in vision–language models applied to automated driving. CARE-Drive compares baseline and reason-augmented model decisions under controlled contextual variation to assess whether human reasons causally influence decision behavior. The framework employs a two-stage evaluation process: prompt calibration to ensure stable outputs, followed by systematic contextual perturbation to measure decision sensitivity to relevant human reasons such as safety margins, social pressure, and efficiency constraints. We demonstrate CARE-Drive in a cyclist overtaking scenario involving competing normative considerations. Results show that explicit human reasons significantly influence model decisions, improving alignment with expert-recommended behavior. However, responsiveness varies across contextual factors, indicating uneven sensitivity to different types of reasons. These findings provide empirical evidence that reason responsiveness in foundation models can be systematically evaluated without modifying model parameters. CARE-Drive operationalizes the tracking requirement of Meaningful Human Control and offers a practical methodology for assessing whether automated decision-making systems exhibit behavior consistent with human-centered reasoning in safety-critical environments.
\end{abstract}


\end{frontmatter}


\newpage

\section{Introduction}
\label{sec:introduction}

Automated driving research is increasingly exploring the use of foundation models, including vision--language models (VLMs) and vision--language--action (VLA) models, for interpreting driving scenes and supporting decision-making. Unlike traditional modular pipelines based on perception, localisation, planning, and control, or end-to-end neural network–based approaches that directly map sensory inputs to driving actions without explicit intermediate reasoning \citep{tampuu2020survey, chen2024end}, foundation models enable unified scene understanding and decision generation through multimodal representations and natural-language interfaces \citep{driess2023palm, zitkovich2023rt, luo2024delving}. Recent studies show that foundation models can assess driving situations, recommend actions, and generate human-interpretable explanations of their decisions \citep{nie2024reason2drive, ishaq2025drivelmm, wang2025alpamayo}. These capabilities make foundation models promising not only as decision-support components but also as evaluators and reasoners for analyzing automated driving behavior \citep{zhou2024vision, gao2025survey}.

Current evaluation methods for automated driving primarily rely on outcome-based metrics, such as collision rate, trajectory displacement error, and rule-based trajectory scoring systems that assess safety compliance and trajectory quality \citep{dauner2024navsim, gariboldi2025vlad}. While these metrics are essential for validating safety and functional correctness, they primarily evaluate the observable outcomes of a maneuver and do not assess whether the underlying decision appropriately reflects human higher-level reasoning, particularly in situations where multiple safe trajectories exist but differ in their alignment with human driving preferences \citep{gariboldi2025vlad, song2025drivecritic}. In ambiguous driving situations involving trade-offs, multiple actions may satisfy safety constraints yet differ in their alignment with human reasoning and normative driving expectations. For example, in an urban driving scenario, an AV might avoid collision with other surrounding vehicles and achieve efficient progress, but can still be perceived as unsafe or unsuitable for human driving \citep{suryana2026reasons}. Consequently, existing evaluation frameworks provide limited insight into whether automated decisions are contextually appropriate from a human-centered perspective, particularly in scenarios where appropriate behavior depends on interpreting and balancing competing situational factors rather than solely achieving safe trajectory outcomes \citep{gariboldi2025vlad, song2025drivecritic}.

Vision--language models have increasingly been applied to automated driving to generate natural-language explanations alongside their decision recommendations, creating the appearance of interpretable and explainable behavior \citep{nie2024reason2drive, ishaq2025drivelmm, gariboldi2025vlad, song2025drivecritic, wang2025alpamayo}. However, prior research has shown that explanations produced by language models may reflect post-hoc rationalizations rather than causal reasoning processes \citep{turpin2023language, parcalabescu2024measuring}. This creates uncertainty about whether stated reasons actually influence the model’s decision, or whether explanations merely provide plausible narrative justifications after the decision has already been determined. Consequently, existing evaluation approaches cannot determine whether model decisions are genuinely responsive to human-relevant considerations, creating uncertainty about the extent to which explanations produced by vision--language models faithfully reflect the underlying decision process \citep{turpin2023language, parcalabescu2024measuring, yeo2025towards}.

Human driving decisions often balance multiple normative considerations, including safety, legality, comfort, and efficiency \citep{michon1985critical, fuller2005towards}. Accordingly, automated driving systems operating in safety-critical environments are increasingly expected to exhibit decision-making behavior that appropriately responds to human-relevant considerations and supports safe and accountable operation \citep{cummings2025identifying}. One framework that provides a principled foundation for designing automated systems to respond to human-relevant considerations is Meaningful Human Control (MHC), which formalizes this requirement through the tracking condition \citep{santoni2018meaningful}. This condition states that automated systems should respond appropriately to the human-relevant reasons that justify a given decision. This perspective is particularly relevant in the context of vision--language models (VLMs), where it remains unclear whether their decisions are genuinely responsive to human-relevant reasons \citep{turpin2023language, parcalabescu2024measuring, song2025drivecritic}. For automated driving systems, satisfying this requirement implies that decisions should appropriately adjust in response to changes in relevant contextual factors, reflecting suitable normative trade-offs \citep{mecacci2020meaningful}. However, existing evaluation frameworks do not provide operational methods to assess whether foundation-model-based decision systems exhibit such reason-responsiveness to human-relevant considerations.

Despite the growing adoption of vision--language models in automated driving, no systematic framework exists to evaluate whether explicit human-centered reasons provided to the model meaningfully influence its decision behavior. In particular, it remains unclear whether normative reasons introduced through prompts measurably affect decision outcomes, or whether they merely alter explanation content without changing the model’s underlying decision preferences. This distinction is critical from the perspective of Meaningful Human Control, which requires that automated system behavior tracks human-relevant reasons rather than only generating plausible post-hoc justifications. Furthermore, it is unknown how model responsiveness to human-relevant reasons varies across different driving contexts involving changes in safety margin, social pressure, and urgency.

This limitation highlights the need for principled evaluation frameworks capable of assessing whether automated decisions genuinely track human-relevant reasons. To address this gap, we propose CARE-Drive (Context-Aware Reasons Evaluation for Driving), a model-agnostic framework for evaluating reason-responsiveness in vision--language models applied to automated driving. CARE-Drive compares baseline model decisions with reason-augmented decisions under controlled contextual variation to assess whether human-centered reasons meaningfully influence observable model decision behavior. By introducing explicit normative reasons through prompt-based specification without modifying model parameters, CARE-Drive enables direct behavioral evaluation of the model’s intrinsic responsiveness to human-centered considerations. The framework isolates prompt-level stability through calibration and evaluates context-sensitive decision changes using systematic variation of observable driving conditions. This approach enables quantitative assessment of alignment between model decisions, human-centered reasons, and expert driving recommendations.

The main contribution of this work are:
\begin{itemize}
    \item \textbf{Framework:} We introduce CARE-Drive, a model-agnostic evaluation framework for measuring reason-responsiveness in vision--language models for automated driving.
    \item \textbf{Methodology:} We propose a two-stage evaluation procedure that isolates prompt-level stability from context-dependent reasoning effects.
    \item \textbf{Empirical analysis:} We conduct a systematic study of how human-centered reasons influence overtaking decisions under controlled contextual variation.
    \item \textbf{Insight:} We demonstrate that explicit normative guidance can shift model decisions toward expert-aligned behavior, while revealing limitations in responsiveness across different contextual factors.
\end{itemize}

\section{Related Work}

\subsection{Vision--Language Models and Evaluation of Automated Driving Decisions}

Recent work has explored the use of vision–language models and reasoning-focused benchmarks for automated driving, including datasets and benchmarks such as Reason2Drive \citep{nie2024reason2drive} and DriveLMM \citep{ishaq2025drivelmm}, as well as multimodal decision-making models such as AlphaMayo \citep{wang2025alpamayo}, VLAD \citep{gariboldi2025vlad}, and DriveCritic \citep{song2025drivecritic}. These works demonstrate that foundation models can interpret traffic scenes, generate structured reasoning, and enable decision prediction and explanation in driving contexts. However, their primary focus is on improving reasoning generation, decision prediction, or explanation fidelity, rather than evaluating whether decisions appropriately respond to explicit human-centered reasons.

In parallel, existing automated driving evaluation frameworks assess decision quality using outcome-based metrics such as collision rate, trajectory accuracy, and rule compliance \citep{dauner2024navsim, gariboldi2025vlad}. While these metrics provide important safety validation, they evaluate observable trajectory outcomes rather than assessing whether decisions appropriately respond to explicit human-centered or normative considerations.

Additionally, prior work has identified limitations in the faithfulness of explanations generated by large language and vision--language models \citep{turpin2023language, parcalabescu2024measuring}. These studies show that explanations may not reliably reflect the underlying decision process. Recent multimodal driving models such as AlphaMayo \citep{wang2025alpamayo} incorporate reasoning supervision by training on human-annotated explanations of driving videos, with the goal of aligning reasoning and decision-making. However, because these explanations are annotated post-hoc based on observed behavior rather than derived from the model’s internal decision process, the training paradigm does not provide direct evidence that the generated reasoning causally influences the model’s decision outcomes. As a result, existing approaches do not provide systematic methodologies for evaluating whether model decisions are genuinely responsive to explicit human-centered reasons.

In contrast, CARE-Drive introduces a framework specifically designed to evaluate reason-responsiveness in vision--language models applied to automated driving.

\subsection{Meaningful Human Control}

Meaningful Human Control (MHC) was originally proposed to address concerns about automated systems whose behavior may lead to harmful outcomes while obscuring responsibility attribution \citep{docherty2015mind}, particularly when human operators are not directly involved in operational decision-making and the system functions as a black box. In such cases, it becomes difficult to establish clear traceability between system behavior and human intent or responsibility. MHC has since been developed as a philosophical and operational framework for ensuring that automated systems remain appropriately aligned with human reasoning and intentions \citep{santoni2018meaningful}.

A central requirement of MHC is the tracking condition, which states that automated system behavior should appropriately track the human-relevant reasons that justify decisions \citep{santoni2018meaningful}, even when humans are not directly involved in operational control. This requirement has been further extended to automated driving systems, where appropriate behavior must reflect human-relevant considerations spanning moral, strategic, tactical, and operational levels of driving decision-making \citep{mecacci2020meaningful}.

Recent work has demonstrated that MHC can be operationalized in automated driving systems with modular architectures by explicitly defining relevant human agents and their associated reasons, and incorporating these factors into formal decision-making frameworks \citep{suryana2025framework, suryana2025human}. In such systems, alignment with human-relevant reasons can be evaluated and enforced by selecting trajectories that best reflect those reasons, thereby preserving traceability between system decisions and human intent. However, these approaches rely on explicit access to intermediate decision representations, which are available in modular pipeline-based systems but not in end-to-end foundation models.

In contrast, end-to-end foundation-model-based systems, including vision--language models, do not expose intermediate decision representations that can be directly aligned with human-relevant reasons. As a result, it remains unclear whether their observable decision behavior appropriately tracks such reasons. CARE-Drive addresses this limitation by providing a model-agnostic evaluation framework that operationalizes the tracking condition behaviorally, by assessing whether model decisions exhibit systematic and interpretable responsiveness to explicit human-relevant reasons. Because CARE-Drive evaluates observable input–output behavior without requiring access to internal model representations, it enables empirical assessment of reason-responsiveness in end-to-end models without requiring modification or retraining of the underlying model.

\section{The CARE-Drive Evaluation Framework}



To operationalize the tracking requirement of Meaningful Human Control in the context of foundation-model-based automated driving, we propose CARE-Drive (Context-Aware Reasons Evaluation for Driving), a model-agnostic evaluation framework for analyzing reason-responsiveness. CARE-Drive enables principled assessment of whether vision–language model (VLM) decisions exhibit observable responsiveness to explicit human-centered reasons in ethically ambiguous scenarios. CARE-Drive is structured as a two-stage framework designed to separate prompt-level stability from context-dependent reasoning effects. In Stage~1 (Prompt Calibration), we identify prompting configurations that produce stable and interpretable VLM outputs by optimizing for consistency and alignment with a reference autonomous-vehicle decision derived from expert reasoning. This step ensures that observed decision shifts are attributable to injected human reasons rather than stochastic prompt variability. In Stage~2 (Contextual Reasons Evaluation), the calibrated prompt is applied under systematic variation of observable scene parameters, such as time-to-collision, rear-vehicle presence, and waiting duration, which correspond to human-relevant normative considerations. By comparing the baseline VLM decision with its reason-augmented counterpart across these controlled perturbations, CARE-Drive evaluates whether and how human-centered reasons causally influence model decision preferences. Observable and systematic responsiveness of model decisions to these injected reasons provides empirical evidence of behavioral tracking, consistent with the Meaningful Human Control framework. This two-stage design enables CARE-Drive to disentangle model instability from genuine context-sensitive reasoning, allowing assessment of both decision alignment and sensitivity to human-relevant trade-offs. An overview of the CARE-Drive framework is shown in Fig.~\ref{fig:CARE-DriveFramework}.

\subsection{Problem Definition}

Let $S = \{(V_i, O_i)\}_{i=1}^N$ represent a set of driving scenes, where $V_i=\{V_{i1}, V_{i2}, \dots, V_{im}\}$ denotes the set of visual representations of driving scenario (e.g., rear-view, bird's-eye view, or other relevant perspectives). The set $O_i=\{O_{i1},O_{i2},\dots,O_{in}\}$ denotes the set of observable contextual variables (e.g., presence and location of surrounding vehicles, time elapsed in a specific driving scenario, and other relevant factors). 
Given a decision instruction $I$, which is a textual part of the prompt representing the decision task, $I$ consists of three  components: $I(R,T,L)$, where $R$ denotes a set of normative human reasons (e.g., safety, fairness), $T$ denotes the thought strategy (e.g., chain-of-thought, tree-of-thought, or no strategy), and $L$ denotes the length of the explanation (e.g., short, long, or unlimited). The decision instruction $I(R,T,L)$ consists of a base role specification $I_{\text{base}}$ that defines the AV’s task (e.g., ‘you are a decision-making component of an automated vehicle’), combined with optional human reasons $R$, a thought strategy $T$, and an explanation-length constraint $L$.

We define the baseline VLM decision as
\[
D^{(0)}_{VLM},E^{(0)}_{VLM} = f_{VLM}(S_i,I(\varnothing,T,L)),
\]
and the reason-augmented decision as
\[
D^{(+R)}_{VLM}, E^{(+R)}_{VLM} = f_{VLM}(S_i,I(R,T,L)).
\]

Here $f_{VLM}$ denotes a vision-language model that maps a driving scene $S$ and a decision instruction $I$ to a recommended decision $D_{VLM}$ and an associated explanation $E_{VLM}$ of length $L$. The difference between the two settings is whether the instruction includes explicit human-centered reason $R$. We assume the existence of a reference decision $D_{AV}(S)$ provided by domain experts, representing the decision that experts suggest should be taken in scene $S$. We study whether and explicitly provided human reasons $R$ shift VLM decision from its baseline output $D_{VLM}^{(0)}$ toward a reference decision $D_{AV}$. 

\subsection{Use Case: Overtaking Scenario}
\label{sec:use_case_overtaking}

To evaluate CARE-Drive, we apply it to a case study involving conflicting human-relevant reasons. By conflicting reasons, we refer to situations in which multiple decision options are available to an automated vehicle, and each option involves trade-offs with respect to human expectations and normative considerations. In this study, the decision space consists of two primary options: (1) overtake the cyclist, or (2) remain behind and follow the cyclist. In CARE-Drive, these normative considerations are explicitly represented as human-relevant reasons $R$ within the decision instruction, enabling controlled evaluation of how such reasons influence model decisions. These normative considerations are operationalized through both explicit human-relevant reasons $R$ provided in the decision instruction and contextual variables $O$, such as time-to-collision with oncoming vehicles, presence of following vehicles, and waiting duration behind the cyclist. This representation enables CARE-Drive to evaluate whether VLM decisions exhibit observable responsiveness to human-relevant reasons under controlled contextual variation.

\begin{figure}
    \centering
    \includegraphics[width=\linewidth]{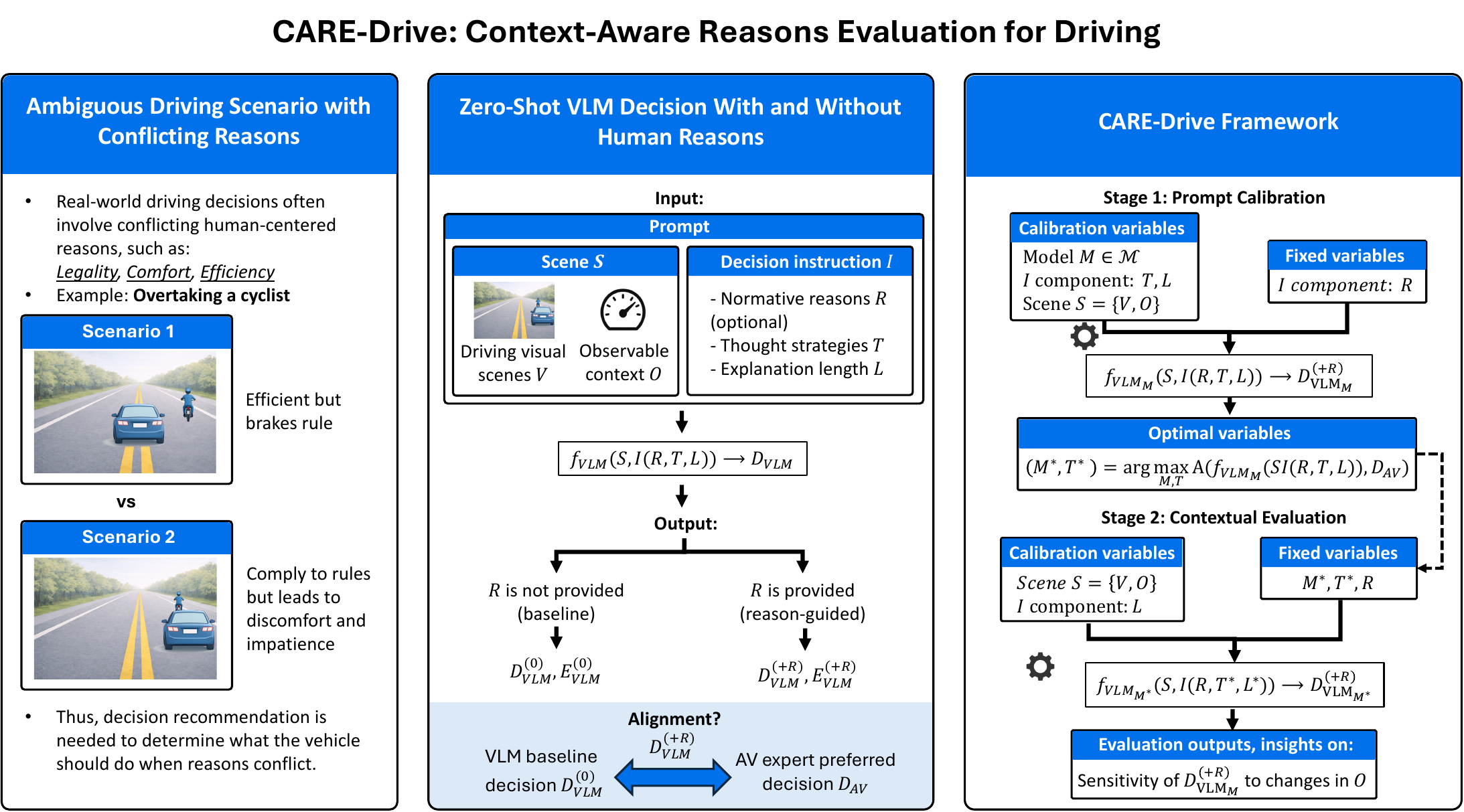}
    \caption{\textbf{Overview of the CARE-Drive framework for evaluating reason-responsive VLM decision-making in ethically ambiguous driving scenarios.}
    \textbf{The left panel} illustrates the motivating setting: similar driving scenes (e.g., cyclist on a no-passing road) can give rise to conflicting reasons (safety, legality, comfort, efficiency), leading to ambiguity whether to overtake.
    \textbf{The middle panel} shows how a vision-language model (VLM) receives a scene representation $S=\{V,O\}$, consisting of a visual input $V$ and observable context $O$, together with a decision instruction $I(R,T,L)$ that specifies injected human reasons $R$, a thought strategy $T$, and an explanation-length regime $L$. The VLM produces a decision $D_{VLM}$ with or without human reasons $(R=\varnothing)$, enabling comparison between baseline and reason-augmented behaviour.
    \textbf{The right panel} depicts the two-stage CARE-Drive evaluation pipeline. In \textit{Stage 1 (Prompt Calibration)}, the model $M$ and thought strategy $T$ are treated as calibration variables. The goal is to identify an optimal configuration $(M^*, T^*)$ that maximises alignment between the reason-augmented VLM decision $V_{VLM}^{(+R)}$ and the expert reference decision $D_{AV}$. In \textit{Stage 2 (Contextual Evaluation)}, the calibrated configuration $(M^*, T^*)$ is held fixed, and the observable context $O$ (e.g., time-to-collision with oncoming vehicles, presence of vehicle behind, passenger urgency) is systematically varied to measure how sensitively VLM augmented with human reasons casually influences decisions across situations.
    \textbf{The bottom arrow in blue} illustrates CARE-Drive's core outcome: quantifying the alignment between baseline VLM decisions, reason-augmented decisions, and expert judgments.}
    \label{fig:CARE-DriveFramework}
\end{figure}

To demonstrate that CARE-Drive can be applied to real-world scenarios, we select a real use case that has previously been used to study the tracking requirement of Meaningful Human Control \citep{suryana2025framework, suryana2025human}, namely overtaking a cyclist. The scenario is adapted from a real-world case \citep{TeslaShort2025}, in which Tesla's Full Self-Driving Beta operates on a bidirectional road and encounters a cyclist ahead. Because the road appears relatively narrow and is separated by double solid yellow lines, indicating that crossing into the opposite lane is prohibited, the automated vehicle chooses to remain behind the cyclist and follow for an extended period of time. Because this scenario involves competing human-relevant reasons that justify different decision options, it provides a suitable testbed for evaluating whether model decisions appropriately track such reasons, consistent with the Meaningful Human Control framework.

Observations from the video indicate that human drivers in similar situations may choose to overtake when it is safe to do so, highlighting the ambiguity inherent in the decision. This ambiguity arises from competing human-relevant considerations. Specifically, overtaking the cyclist may improve efficiency and reduce discomfort experienced by the cyclist due to prolonged following, while remaining behind the cyclist ensures compliance with legal constraints but reduces efficiency and may increase discomfort due to extended travel time. Importantly, the opposite lane is visibly unoccupied, further contributing to the plausibility of overtaking despite the legal restriction. These competing considerations illustrate a trade-off among legality, efficiency, and comfort, making the scenario suitable for evaluating whether automated decisions appropriately track human-relevant reasons. This scenario therefore enables systematic application of CARE-Drive by comparing baseline and reason-augmented VLM decisions across controlled contextual variations. An illustration of this scenario is shown in the left panel of Figure~\ref{fig:CARE-DriveFramework}.

\subsection{Stage 1: Prompt Calibration}

Stage~1 identifies a prompt configuration that yields stable, interpretable, and expert-aligned decisions before any context-sensitivity analysis is performed. 
The purpose of this stage is to isolate prompt-level effects—model choice and reasoning strategy—from contextual variation, ensuring that subsequent decision changes can be attributed to observable context rather than prompt instability.

During Stage~1, we vary the following prompt-level parameters:
\begin{itemize}
    \item \textbf{Model $M$}: the vision--language model used to generate the driving decision.
    \item \textbf{Thought strategy $T$}: the reasoning structure imposed on the model (No-Thought, Chain-of-Thought, or Tree-of-Thought).
\end{itemize}

All other elements of the prompt, including the injected human reasons $R$, the scene representation $S=\{V,O\}$, and decoding parameters (temperature and top-$p$), are held fixed unless explicitly stated.

The goal of Stage~1 is to identify an optimal prompt configuration $(M^*,T^*)$ that produces reason-augmented decisions $D^{(+R)}_{\mathrm{VLM}}$ which are both consistent across repeated stochastic runs and aligned with the expert reference decision $D_{AV}$.

The expert reference decision is drawn from \citep{suryana2026reasons}, which reports that, in the overtaking scenario described in Section~\ref{sec:use_case_overtaking}, a clear majority of AV experts recommend overtaking the cyclist despite the presence of a legal prohibition. This recommendation reflects an explicit trade-off between safety, efficiency, comfort, and legal compliance.

\subsubsection{Calibration parameters}

To ensure that the parameters $M$ and $T$ produce output that is both consistent and aligned with the reference decision, we perform a calibration process. This involves adjusting the values of $M$ and $T$ until the output meets the desired criteria. In addition to these parameters, the explanation length $L$, visual scene $V$ and observable context $O$ is also modified, but they are not calibrated in this stage, as they will serve as test parameters to evaluate sensitivity in Stage 2. The following outlines the types of parameter values we will test to identify those that yield consistent and aligned outputs with the reference decision:
\\

\noindent \textbf{Visual Scene $S$ and Observable Context $O$} \\
We define three scenarios, each consisting of different combinations of visual scenes and observable contexts. Each scenario replicates the situation described in Section~\ref{sec:use_case_overtaking}, with systematic variation in the visual scene and observable context based on situations that AV experts describe as relevant for driving decisions.

\begin{itemize}
    \item \textbf{Scenario 1 -- Baseline Scene}: No oncoming vehicles and no vehicles following the AV.
    \item \textbf{Scenario 2 -- Oncoming}: One oncoming vehicle, with no following vehicles behind the AV.
    \item \textbf{Scenario 3 -- Behind}: A vehicle following the AV, but no oncoming vehicles.
\end{itemize}

In this representation, the scene $S=\{V,O\}$ is constructed as follows. 
The visual scene $V$ is given by a static image of the driving situation from the driver’s perspective. Because rear vehicles are not visible in the dashboard view, Scenario~1 (baseline) and Scenario~3 (vehicle behind) share the same visual scene, while Scenario~2 includes an oncoming vehicle that is visible in the image.

\begin{figure}
    \centering
    \includegraphics[width=0.9\linewidth]{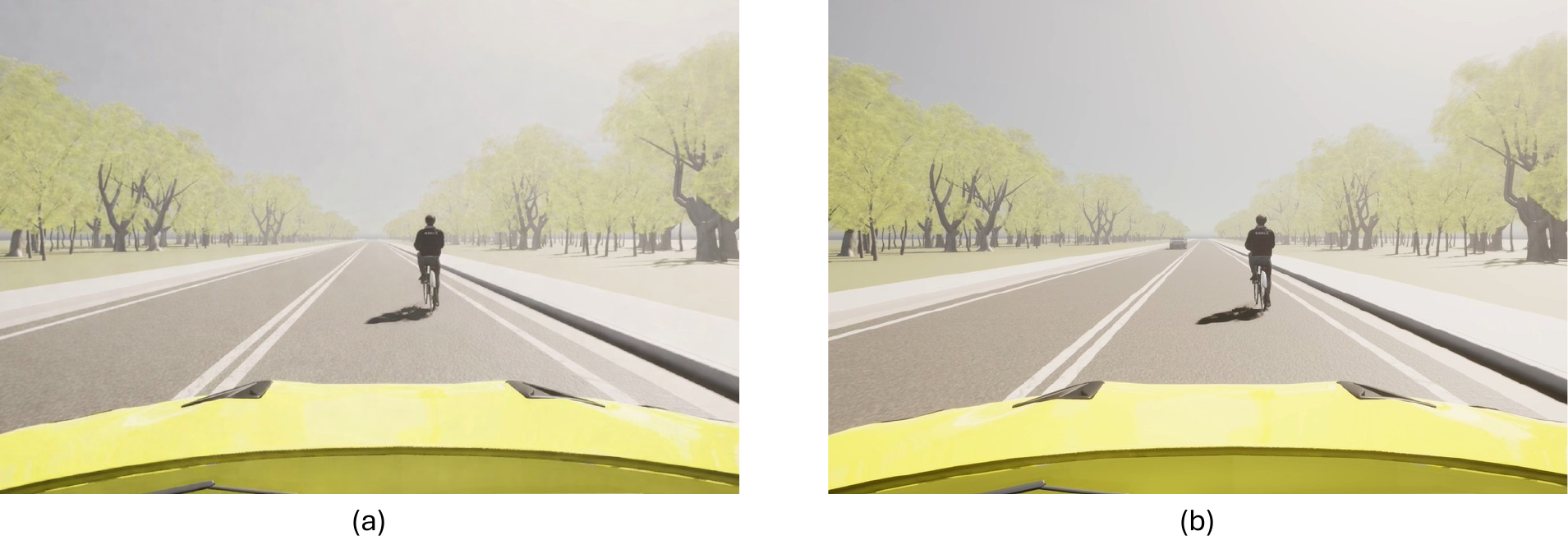}
    \caption{
Visual scenes used in CARE-Drive. The scenes are taken from CARLA simulator. (a) Scenario~1 (baseline) and Scenario~3 (vehicle behind) share the same dashboard image, since rear vehicles are not visible in $V$. 
(b) Scenario~2 includes an oncoming vehicle visible in the image. 
The vehicle behind in Scenario~3 is represented only in the observable context $O$. 
}

    \label{fig:placeholder}
\end{figure}

The observable context $O$ is defined by numerical and categorical variables describing the same scene, including cyclist speed, AV speed, speed limit, longitudinal distance to the cyclist, and the presence or absence of oncoming traffic and vehicles following the AV. While the exact speed of the cyclist is unknown, an average speed of 13~km/h is assumed, based on data from a study on predicting cyclist speeds~\citep{average-cyclist}. The AV speed is below the typical 50~km/h speed limit for rural roads where cyclists are permitted. The Dutch guideline for driving behind another vehicle recommends a two-second following distance~\citep{verkeersregel}, which corresponds to a minimum longitudinal distance of 7.3~m between the cyclist and the vehicle. For safety and realism, this distance is rounded up to 10~m. The longitudinal distance, speed limit, cyclist speed, and AV velocity are kept constant across all three scenarios. Although Scenario~1 and Scenario~3 have identical visual scenes, their difference is encoded in $O$ through the presence or absence of a vehicle behind the AV.

In addition, $O$ includes fixed geometric and kinematic parameters used in the prompt, including lane width (3~m), vehicle width (1.8~m), lateral offset of the cyclist from the lane centerline (2.5~m), and the assumption that the AV drives centered within its lane. \\

\noindent \textbf{Thought Strategy $T$} \\
In CARE-Drive, the thought strategy $T$ specifies the reasoning template used within the decision instruction $I(R,T,L)$. We evaluate three strategies, 
$T \in \{\text{No-Thought}, \text{Chain-of-Thought}, \text{Tree-of-Thought}\}$, while keeping the scene description $S=\{V,O\}$ and the injected reasons $R$ fixed. 
When a thought strategy is enabled, the model is explicitly instructed to follow the corresponding reasoning structure (i.e., the structure is enforced via the prompt).

\begin{itemize}
    \item \textbf{No-Thought}: The model is directly asked to produce a decision (without an explicit intermediate reasoning procedure).
    \item \textbf{Chain-of-Thought (CoT)}: The model is instructed to reason step by step before producing the decision \citep{weiCoT}. CoT is widely used and often reported to improve multi-step reasoning performance \citep{PromptEngineering}.
    \item \textbf{Tree-of-Thought (ToT)}: The model is instructed to explore multiple reasoning branches before selecting a decision \citep{yaoToT}. In our implementation, ToT explicitly evaluates two main branches (stay behind vs.\ overtake) and their associated benefits and risks.
\end{itemize}

The thought strategies differ only in the reasoning structure imposed on the model; the underlying scene description and available decision options remain identical across thought strategies.
\\

\noindent \textbf{Explanation Length $L$} \\
The explanation length $L$ controls how much justification $E_{VLM}$ the model is allowed to produce alongside the decision $D_{VLM}$. In all conditions, the decision is always reported in the same discrete format (Decision: case~1 or case~2); only the amount of justification is constrained.

We evaluate two explanation-length regimes:
$L \in \{\text{Few-Sentences}, \text{No-limit}\}$.
In the Few-Sentences condition, the model must provide a brief justification, while in the No-limit condition the model is allowed to produce an unconstrained explanation.

The length constraint is enforced via prompt instructions rather than post-hoc truncation. Although a decision-only output was initially considered, we exclude it from the evaluation because it provides no access to the model’s internal reasoning, which is required to determine whether decisions are influenced by the injected human reasons rather than by stochastic variation. Across all $L$ conditions, the decision is always reported in the same discrete format (Decision: case~1 or case~2); only the justification is constrained.

The Few-Sentences condition does not impose a strict token limit, as our objective is not to constrain surface length but to induce a qualitatively concise reasoning regime. Instruction-based control allows the model to allocate space adaptively, rather than being artificially cut off. In practice, this produces explanations that are substantially shorter than unconstrained outputs (approximately five-fold on average; see Results).

Because chain-of-thought and tree-of-thought strategies rely on the availability of reasoning space, limiting $L$ directly constrains how much structured reasoning can be expressed. This allows us to study whether decision alignment and consistency depend on the availability of extended explanations rather than on the decision format itself. Prior work suggests that excessive verbosity can reduce precision \citep{nayab2024concise}, motivating the inclusion of constrained explanation regimes in our evaluation.
\\

\noindent \textbf{Model $M$} \\
CARE-Drive is model-agnostic and can be applied to any vision–language model. In this study, we evaluate three GPT-based vision–language models that are routinely used in our laboratory for VLM-based driving experiments. Specifically, we select $M \in \{\text{gpt-4.1}, \text{gpt-4.1-mini}, \text{gpt-4.1-nano}\}$, which differ primarily in model capacity and computational efficiency rather than in their overall modeling paradigm.

All models receive identical inputs, including the same scene representation $S=\{V,O\}$ and the same decision instruction $I(R,T,L)$. Decoding parameters (temperature and top-$p$) are held fixed across models to ensure comparable stochasticity. For each combination of model, thought strategy, and reason condition, we perform 30 independent runs, resulting in 12 experimental configurations in total.

Although newer models (e.g., GPT-5.2) were available at the time of writing, they were excluded due to access and cost constraints. Because CARE-Drive is model-agnostic, future users can directly apply the framework to newer models without changing the evaluation procedure.
\\

\noindent \textbf{Human Reasons $R$} \\
Human reasons are defined as normative considerations that govern how an AV should evaluate and prioritize different aspects of a driving maneuver. We adapt a structured set of 13 reason categories from \citep{suryana2026reasons}, derived from interviews with AV experts across multiple disciplines. These include safety, rule compliance, efficiency, comfort, environmental impact, social appropriateness, fairness, cultural adaptation, acceptance, interaction, vigilance and readiness, continuous control, and control transition.

In CARE-Drive, $R$ is provided to the model as a structured normative policy in which avoiding physical harm is always prioritized, and the remaining reasons (e.g., efficiency, legality, comfort, and fairness) are weighed against each other using explicit principles when conflicts arise. These reasons are scenario-independent and do not encode scene-specific facts.

All 13 reasons are always included; we do not vary subsets of reasons, because CARE-Drive evaluates whether the presence of a full normative specification influences the model’s decision. The reasons are outcome-neutral and do not encode the expert’s final decision, but only the considerations experts report using to justify their recommendations.

We compare the baseline model output $D^{(0)}_{VLM}$ with the reason-augmented output $D^{(+R)}_{VLM}$ to assess whether explicit normative guidance shifts VLM decisions toward expert judgments. This approach is consistent with prior work showing that instance-specific guiding signals can significantly improve LLM reasoning accuracy \citep{li2023guiding}. \\

\noindent \textbf{Instruction Structure $I$} \\
Here, every instruction includes a fixed base task specification $I_{\text{base}}$ that defines the model’s role as an automated-vehicle decision maker (e.g., ``you are a decision-making component of an AV; decide what the vehicle should do''). Thus, the full instruction used in all conditions is
\[
I = I_{\text{base}} \cup I(R,T,L).
\]
In the experimental tables, this base component is referred to as \textbf{Role}, thus $I_{\text{base}}=\textbf{Role}$. Accordingly, ``Role-only'' corresponds to $I_{\text{base}}$ with $R=\varnothing$, while ``Role+CoT'' and ``Role+ToT'' correspond to $I_{\text{base}} \cup I(\varnothing,T,L)$, and their reason-augmented counterparts include $R$. 

\subsubsection{Finding optimal parameters}
\label{sec:FindingOptimalParameters}

The optimal parameter configuration is determined in two steps for the following experimental setup: the scene $S$ is fixed to \textbf{Scenario 1 -- Baseline Scene}, and the length of explanation length is fixed $L=\textbf{No-Limit}$. \\

\noindent \textbf{Step 1: Identifying the optimal model $M$ and thought strategy $T$.} \\ In Step~1, we perform a calibration procedure to identify the model $M$ and thought strategy $T$ that most consistently reproduce the expert reference decision under a fixed and normatively salient driving context. The expert reference decision $D_{AV}$ in this setting is binary (overtake vs.\ stay behind). Prior expert elicitation indicates that, for the baseline scenario considered here, most experts recommend overtaking. Accordingly, alignment is evaluated by counting how often the model produces the same decision as the expert reference.

For each candidate configuration $(M,T)$, we run the reason-augmented VLM
\[
D^{(+R)}_{VLM_M} = f_{VLM_M}(S, I(R,T,L))
\]
for $N = 30$ independent stochastic runs under identical conditions. Alignment is measured as the number of runs in which the model’s decision matches the expert reference decision:
\[
C(M,T) = \sum_{i=1}^{N}
\mathbbm{1}\!\left[
D^{(+R,i)}_{VLM_M}(S, I(R,T,L)) = D_{AV}(S)
\right],
\]
where $\mathbbm{1}[\cdot]$ denotes the indicator function and $i$ indexes repeated runs of the same configuration.

During Step~1, the explanation length $L$ is fixed to the \textit{No-Limit} setting in order to avoid constraining the model’s reasoning capacity during calibration. This ensures that differences in alignment can be attributed to the model $M$ and the imposed reasoning structure $T$, rather than to limitations on explanation length.

The observable context is also fixed to a following time of 24~seconds, which exceeds both the cyclist discomfort threshold (10~seconds) and the driver impatience threshold (15~seconds). This choice intentionally activates normative tension between legality, comfort, and efficiency. These thresholds are grounded in empirical findings showing that prolonged close following increases discomfort and perceived risk for cyclists~\citep{oskina2023safety}, and that extended delays behind slow road users increase driver frustration and the tendency to consider overtaking~\citep{kinnear2015experimental}.

In addition, we require that the model explicitly acknowledges that overtaking violates traffic rules, ensuring that any decision to overtake reflects a deliberate normative trade-off rather than ignorance of legality.

The optimal configuration $(M^*,T^*)$ is selected as the one that maximizes the alignment count $C(M,T)$ over the 30 runs. This step serves as a screening procedure to identify stable and interpretable prompt configurations before conducting the contextual sensitivity analysis in Stage~2. \\

\noindent \textbf{Step 2: Evaluating robustness across context and explanation length.} \\ 
In Step~2, we evaluate the robustness of the calibrated configuration $(M^*,T^*)$ across three qualitatively distinct driving scenarios that differ in visual scene $V$ and observable context $O$, as well as under different explanation-length regime $L$.

In the implementation, scenario differences are represented by predefined visual inputs $V_s$ and fixed observable context assignments $O_s$, rather than by parametric variation of context variables. Specifically, we define a finite set of scenarios , $s \in \mathcal{S} = \{1,2,3\}$,
where each scenario corresponds to a fixed scene instance $(V_s,O_s)$ representing a qualitatively distinct driving situation (baseline, oncoming vehicle, and vehicle behind).

\begin{figure}
    \centering
    \includegraphics[width=\linewidth]{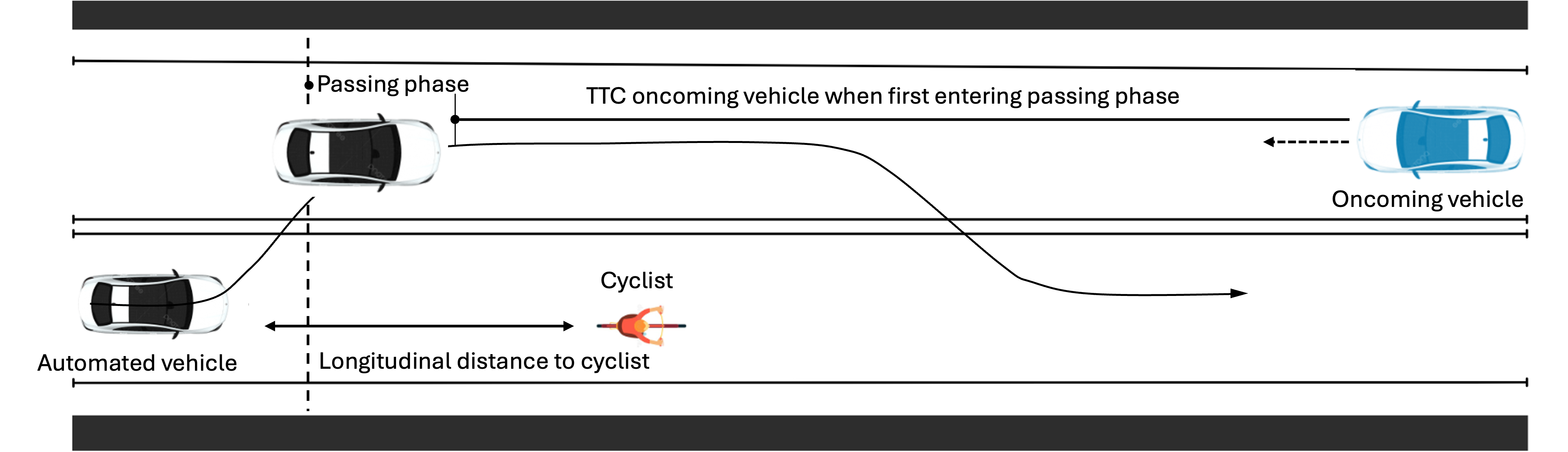}
    \caption{Illustration of the overtaking scenario with an oncoming vehicle. The passing phase, defined as the moment when the ego vehicle is in the opposing lane while overtaking the cyclist, is used as the reference point for calculating the time-to-collision (TTC) with the oncoming vehicle. The longitudinal distance to the cyclist is defined when the ego vehicle is directly behind the cyclist prior to initiating the overtaking maneuver.}
    \label{fig:overtaking_figure}
\end{figure}

Consistent with the scene definition $S=\{(V_i,O_i)\}_{i=1}^N$ where $V_i=\{V_{i1},\dots,V_{im}\}$, the visual input for each scenario is defined as a set of one or more visual observations:
\[
V_s =
\begin{cases}
\{V_{s1}\}, & s=1, \\
\{V_{s1}, V_{s2}\}, & s=2, \\
\{V_{s1}\}, & s=3,
\end{cases}
\]
where $V_{sj}$ denotes the $j$-th visual observation of scenario $s$. The visual representations of the scenarios are shown in Fig.~\ref{fig:placeholder}. Scenario~2 includes two visual observations corresponding to the presence of an oncoming vehicle, as illustrated in Fig.~\ref{fig:placeholder}(b) and Fig.~\ref{fig:overtaking_figure}, which together capture the sequential visual context relevant to the overtaking decision. In contrast, Scenario~1 and Scenario~3 each include a single visual observation, shown in Fig.~\ref{fig:placeholder}(a), respectively. Although Scenario~1 and Scenario~3 share the same visual input, they differ in observable context, reflecting the absence or presence of a vehicle behind the ego vehicle.

The observable context tuple for each scenario is defined as a fixed assignment $O_s = (TTC_{o,s}, B_s, F_s)$, with scenario-specific values given by
\[
O_s =
\begin{cases}
(\varnothing,\ 0,\ 24), & s=1, \\
(8.5,\ 0,\ 24), & s=2, \\
(\varnothing,\ 1,\ 24), & s=3,
\end{cases}
\]
where $TTC_{o,s}$ denotes the time-to-collision with an oncoming vehicle, $B_s \in \{0,1\}$ indicates the presence of a vehicle behind the ego vehicle, and $F_s$ denotes following time in seconds. The symbol $\varnothing$ indicates that no oncoming vehicle is present in the scene.

The explanation-length regime is varied independently as $L \in \{0,1\}$, where $L=0$ corresponds to the No-limit condition and $L=1$ corresponds to the Few-Sentences condition.

The complete experimental condition evaluated in Step~2 is therefore defined as
\[
(V,O,L) = (V_s, O_s, L),
\quad s \in \mathcal{S},\ L \in \{0,1\}.
\]

Step~2 evaluates robustness across these three predefined scene instances. Unlike Stage~2, which performs a full-factorial sensitivity analysis over context variables, Step~2 does not vary the numerical values of $(TTC_o,B,F)$ but instead verifies that the calibrated configuration $(M^,T^)$ produces stable and interpretable decisions across qualitatively distinct scenario types. The detailed fixed scenario assignments are summarized in Table~\ref{tab:scenario_spec}. The selected value of $TTC_o = 8.5$~s represents a conservative safety margin relative to typical human overtaking behavior, as naturalistic driving data show that drivers complete overtaking maneuvers in the presence of oncoming vehicles at mean TTC values of approximately $3.30 \pm 1.70$~s \citep{kovaceva2019drivers}.

In addition to safety considerations, prolonged following of cyclists has been shown to influence both cyclist comfort and driver overtaking behavior. Experimental studies demonstrate that cyclists perceive increased risk and discomfort when being followed for durations of approximately 20 seconds \citep{oskina2023safety}. Furthermore, empirical analyses of overtaking dynamics show that accumulated delay affects overtaking behavior, with measurable behavioral adaptations occurring over delay intervals on the order of tens of seconds \citep{llorca2013influence}. Based on these findings, a following duration of 24 seconds was selected to represent a realistic and empirically grounded delay level at which efficiency-related motivations become behaviorally relevant. This enables evaluation of whether the calibrated VLM configuration exhibits sensitivity to delay-based efficiency considerations while maintaining safety-consistent decision-making.

\begin{table}[h]
\centering
\caption{Fixed scene instances used in Step~2 robustness evaluation. Each scenario corresponds to a predefined scene $(V_s,O_s)$ rather than a parametric variation of context variables.}
\label{tab:scenario_spec}
\begin{tabular}{llll}
\hline
\textbf{Component} & \textbf{Scenario 1} & \textbf{Scenario 2} & \textbf{Scenario 3} \\
\hline
Visual input $V_s$ & $\{V_{s1}\}$ & $\{V_{s1},V_{s2}\}$ & $\{V_{s1}\}$ \\
$TTC_o$ (s) & $\varnothing$ & $8.5$ & $\varnothing$ \\
Vehicle behind $B$ & $0$ & $0$ & $1$ \\
Following time $F$ (s) & $24$ & $24$ & $24$ \\
\hline
\end{tabular}
\end{table}

For each unique combination of $(V,O,L)$, the VLM decision
\[
D^{(+R)}_{VLM_{M^*}}(V,O,L)
\]
is evaluated over $N = 30$ independent runs using the GPT-4.1 vision-language model. Using the same binary indicator definition as in Step~1, we define
\[
Y = \mathbbm{1}\!\left[
D^{(+R)}_{VLM_{M^*}}(V,O,L) = \text{Overtake}
\right],
\]
where $Y=1$ denotes an overtaking decision and $Y=0$ denotes staying behind.

The empirical overtaking rate
\[
P(Y=1)
\]
is estimated as the proportion of runs in which $Y=1$.

This step evaluates whether a prompt configuration that aligns with expert judgment in the calibrated baseline scenario remains stable and context-sensitive when observable driving conditions and reasoning bandwidth vary across scenario-specific visual and contextual configurations. In contrast to Step~1, which focuses on alignment under fixed conditions, Step~2 assesses robustness and generalization across qualitatively distinct scenario types.

\subsection{Stage 2: Contextual Evaluation}

In Stage~2, we evaluate how sensitive the calibrated CARE-Drive decision process is to variations in observable driving context. The goal of this stage is to isolate how changes in situational variables influence the VLM’s decision once prompt-level instability has been removed.

All experiments in this stage use the fixed configuration $(M^*,T^*)$ obtained in Stage~1. The visual scene $V$ is also held fixed and corresponds to Scenario~2 (oncoming vehicle), since overtaking risk is only meaningful in the presence of opposing traffic. All variation in this stage therefore occurs through the observable context $O$ and the explanation length $L$.

\subsubsection{Context sensitivity analysis} 

To evaluate how variations in driving context influence VLM decision-making, we conduct a full-factorial analysis in which the observable context $O$ and the explanation-length constraint $L$ are systematically varied. All experiments in this section use the calibrated configuration $(M^*,T^*)$ obtained in Stage~1 and a fixed visual scene $V$ corresponding to Scenario~2 (oncoming vehicle). Scenario~2 is selected because it contains an explicit safety trade-off involving oncoming traffic. This allows systematic evaluation of whether the VLM’s decisions respond appropriately to variations in safety-critical context while holding the visual scene constant.

The observable-context variables considered in this analysis are summarized in Table~\ref{tab:variables}. Together, these variables define the context tuple
\[
O = (TTC_o, B, U, F),
\]
where each variable captures a factor that human drivers and AV experts identify as relevant for overtaking decisions. In addition to these context variables, we vary the explanation-length regime $L$ as a controlled prompt-level parameter. Although $L$ is not an observable traffic variable, it directly constrains the model’s available reasoning bandwidth and affects response latency, which is critical for real-time AV deployment.

\begin{table}[H]
\centering
\caption{Varied variables used in Stage~2 to evaluate $D^{(+R)}_{VLM}$ under variations of the baseline oncoming-vehicle scenario (Scenario~2).}
\label{tab:variables}
\begin{tabular}{|c|p{5.2cm}|p{8.5cm}|}
\hline
\textbf{\#} & \textbf{Varied variable} & \textbf{Symbolic definition} \\
\hline
1 & Time-to-collision with oncoming vehicle &
$TTC_o \in \{1.7,\,3.4,\,5.1,\,6.8,\,8.5\}\;\text{s}$ \\

2 & Vehicle behind indicator &
$B \in \{0,1\}$, where $B=1$ indicates a vehicle following the AV \\

3 & Passenger urgency indicator &
$U \in \{0,1\}$, where $U=1$ denotes a passenger in a hurry \\

4 & Following time behind cyclist &
$F \in \{12,\,18,\,24\}\;\text{s}$ \\

5 & Explanation-length regime &
$L \in \{0,1\}$, where $L=1$ indicates limit on explanation length \\

\hline
\end{tabular}
\end{table}

Given this configuration, we define a binary outcome variable
\[
Y = \mathbbm{1}\!\left[D^{(+R)}_{VLM_{M^*}}(V,O,L) = \text{Overtake}\right],
\]
where $\mathbbm{1}[\cdot]$ denotes the indicator function. Thus, $Y=1$ corresponds to an overtaking decision and $Y=0$ corresponds to staying behind the cyclist.

For each unique combination of $(TTC_o, B, U, F, L)$, the VLM is evaluated over 30 independent runs. The empirical overtaking probability
\[
P(Y=1)
\]
is estimated as the proportion of runs in which $Y=1$.

The selected variables reflect established determinants of human overtaking behavior. Time-to-collision with oncoming traffic is a primary safety cue in overtaking decisions \citep{piccinini2018influence}. The presence of a vehicle behind the AV introduces social pressure and risk, as emphasized by AV experts \citep{suryana2026reasons}. Passenger urgency is included because verbalized time pressure can alter driving behavior and AV decision-making \citep{talk2drive}. Following time behind the cyclist is varied because prolonged close following increases cyclist discomfort and perceived risk \citep{oskina2023safety}. Finally, explanation length $L$ is varied to assess how reasoning bandwidth affects decision stability and latency.

Vehicle-behind and passenger-urgency variables are encoded exclusively in the observable context $O$ rather than in the visual scene $V$, since rear vehicles and passenger states are not visible in the dashboard-view image.

\subsubsection{Binary logit method}

Each experimental condition produces a binary decision
$Y \in \{0,1\}$, where $Y=0$ denotes staying behind the cyclist and $Y=1$ denotes overtaking.
To quantify how observable context influences this decision, we estimate a binary logistic regression model.

The log-odds of overtaking are defined as:

\begin{equation}
\text{logit}(p) = \ln\left(\frac{p}{1-p}\right)
= \beta_0
+ \beta_1 \text{$TTC_o$}
+ \beta_2 \text{$B$}
+ \beta_3 \text{$U$}
+ \beta_4 \text{$F$}
+ \beta_5 \text{$L$},
\end{equation}

where $p = P(Y=1)$ is the probability of overtaking, and $\beta_0$ represents the intercept corresponding to baseline log-odds when all predictors are zero. To obtain the overtaking probability directly, the inverse logit transformation is applied:

\begin{equation}
p =
\frac{1}{1 + \exp\left(
-(\beta_0
+ \beta_1 \text{$TTC_o$}
+ \beta_2 \text{$B$}
+ \beta_3 \text{$U$}
+ \beta_4 \text{$F$}
+ \beta_5 \text{$L$})
\right)}.
\end{equation}

This transformation maps log-odds to probability in the range $[0,1]$. Changes in predictor values modify the log-odds linearly but influence overtaking probability nonlinearly. To facilitate interpretation, predicted probabilities are computed under baseline conditions and after varying individual predictors while holding other variables constant.

Categorical variables ($B$, $U$, and $L$) are encoded as binary indicators. Continuous variables ($TTC_o$ and $F$) are normalized to the range $[0,1]$ to ensure consistent scaling and comparability across predictors.

\subsubsection{Odds ratio and probability interpretation}

To quantify the influence of each predictor on overtaking behavior, odds ratios are computed as:

\begin{equation}
\text{Odds Ratio} = e^{\beta_i}.
\end{equation}

An odds ratio greater than 1 indicates that the predictor increases the odds of overtaking, while a value less than 1 indicates a reduction in overtaking tendency. While odds ratios provide a multiplicative measure of effect size, predicted probabilities are additionally computed to provide intuitive interpretation of behavioral impact.

Specifically, the baseline overtaking probability is obtained from the intercept term:

\begin{equation}
p_{\text{baseline}} =
\frac{1}{1 + e^{-\beta_0}},
\end{equation}

and the probability under modified conditions is computed by incorporating the corresponding predictor values into the inverse logit function. This allows direct quantification of how contextual factors increase or decrease overtaking likelihood.

Together, the logit coefficients, odds ratios, and predicted probabilities provide complementary perspectives: coefficients quantify directional influence, odds ratios measure effect magnitude, and probabilities provide interpretable behavioral likelihood.

\subsubsection{CARLA validation}

Finally, selected contextual conditions are replayed in the CARLA simulator to validate that the learned decision boundaries correspond to physically feasible overtaking maneuvers. CARLA is used for behavioral validation rather than statistical estimation, ensuring that the context-sensitive decisions identified by CARE-Drive translate into executable AV behavior.

\section{Results}

\subsection{Stage-1 calibration results}
As defined in the CARE-Drive framework, the goal of Stage~1 is to identify the model $M$ and thought strategy $T$ that yield the most stable and expert-aligned decisions under a fixed, normatively challenging driving situation. This calibration stage isolates prompt-level effects by keeping the scene $S$ and observable context $O$ constant while varying the prompt parameters. The outcome of this stage is the optimal configuration $(M^*,T^*)$.

\subsubsection{Baseline vs.\ effect of human reasons}
We first test whether explicitly providing human reasons $R$ changes the VLM’s decision relative to its baseline behavior. The baseline condition corresponds to
\[
D^{(0)}_{VLM} = f_{VLM}(S, I(\varnothing, T, L)),
\]
where no normative reasons are injected. In contrast, the reason-augmented condition is
\[
D^{(+R)}_{VLM} = f_{VLM}(S, I(R, T, L)).
\]

Across all tested combinations of model $M$ and thought strategy $T$, the baseline condition $R=\varnothing$ produced an overtaking rate of $0\%$ over 30 runs under identical temperature and top-p: the VLM always chose to stay behind the cyclist. This held for all three prompt structures (Role-only, Role+CoT, and Role+ToT) and for all three models. In other words, without explicit human reasons, the VLM systematically defaulted to strict rule compliance.

By contrast, when the same prompts were augmented with human reasons ($R\neq\varnothing$), overtaking behavior emerged and varied substantially across $M$ and $T$. These results are summarized in Table \ref{tab:stage1_calibration}. The presence of $R$ therefore induces a clear shift from the baseline decision,
\[
D^{(0)}_{VLM} \;\neq\; D^{(+R)}_{VLM},
\]
demonstrating that explicit normative guidance is necessary for the model to reproduce the expert-recommended behavior in this scenario.

\begin{table}[H]
\centering
\caption{Overtaking counts in the baseline calibration scenario (24\,s following time) for different prompt configurations and models. Each condition was run 30 times.}
\label{tab:stage1_calibration}
\begin{tabular}{l c c}
\hline
\textbf{Prompt configuration} & \textbf{Model $M$} & \textbf{Overtake count} \\
\hline
Role                           & gpt-4.1          & 0  \\
Role                           & gpt-4.1-mini     & 0  \\
Role                           & gpt-4.1-nano     & 0  \\
\hline
Role + R                      & gpt-4.1          & 20 \\
Role + R                      & gpt-4.1-mini     & 15 \\
Role + R                      & gpt-4.1-nano     & 0  \\
\hline
Role + R + CoT                & gpt-4.1          & 30 \\
Role + R + CoT                & gpt-4.1-mini     & 28 \\
Role + R + CoT                & gpt-4.1-nano     & 8  \\
\hline
Role + R + ToT                & gpt-4.1          & 30 \\
Role + R + ToT                & gpt-4.1-mini     & 25 \\
Role + R + ToT                & gpt-4.1-nano     & 11 \\
\hline
\end{tabular}
\end{table}

\subsubsection{Step 1: Screening for viable $(M,T)$ under a fixed high-tension baseline}
\label{sec:stage1_step1}

In the first step, we screen candidate model--reasoning-strategy pairs $(M,T)$ under a fixed, high-tension baseline condition. 
The scene, observable context, and explanation length are held constant to isolate whether the model can follow human reasons at all.

Specifically, we fix Scenario~1 (no oncoming traffic, no vehicle behind), a following time of $F=24$\,s, and $L=\textit{Few-Sentences}$, which jointly activate normative conflict between legality, comfort, and efficiency.
Each $(M,T)$ combination is evaluated over 30 stochastic runs, and alignment is measured as the frequency with which the reason-augmented decision $D^{(+R)}_{\mathrm{VLM}}$ matches the expert reference decision $D_{AV}$.

This step functions as a normative capability filter: only configurations that reliably follow expert-aligned human reasons are allowed to proceed to robustness evaluation.

As shown in Table~\ref{tab:stage1_calibration}, only \texttt{gpt-4.1} with Chain-of-Thought (CoT) and Tree-of-Thought (ToT) achieve perfect alignment under these conditions. 
All other configurations are excluded from further consideration.

\subsubsection{Step 2: Robustness across scenarios and explanation length}
\label{sec:stage1_step2}

The surviving candidates from Step~1 are now subjected to a robustness evaluation.
While Step~1 establishes whether a configuration can follow human reasons at all, Step~2 determines which configuration remains aligned when context and explanation conditions vary. We therefore fix the model to \texttt{gpt-4.1} and compare CoT and ToT across all three scenarios (Scenarios~1--3) and across both explanation-length regimes $L\in\{\text{Few-Sentences},\text{No-Limit}\}$.
Each condition is evaluated over 30 stochastic runs.



\begin{table}[t]
\centering
\caption{Stage~1 (Step~2) calibration refinement: robustness across scenarios and explanation-length regimes for \texttt{gpt-4.1}. Values report average generation time (s) and overtaking rate (\%) over 30 runs per condition.}
\label{tab:calibration_step2_robustness}
\setlength{\tabcolsep}{6pt}
\renewcommand{\arraystretch}{1.15}
\begin{tabular}{llcccc}
\toprule
\textbf{Scenario} & \textbf{Prompt} &
\multicolumn{2}{c}{\textbf{No-limit}} &
\multicolumn{2}{c}{\textbf{Few-sentences}} \\
\cmidrule(lr){3-4} \cmidrule(lr){5-6}
& & \textbf{Time (s)} & \textbf{Overtake (\%)} & \textbf{Time (s)} & \textbf{Overtake (\%)} \\
\midrule
Scenario~1 (None)   & Role + HR + CoT & 18.037 & 50.00  & 3.950 & 96.67 \\
                   & Role + HR + ToT & 16.600 & 76.67  & 4.468 & 96.67 \\
\addlinespace
Scenario~2 (Oncoming) & Role + HR + CoT & 18.000 & 30.00  & 5.221 & 0.00 \\
                      & Role + HR + ToT & 19.568 & 93.33  & 5.630 & 0.00 \\
\addlinespace
Scenario~3 (Follow) & Role + HR + CoT & 19.900 & 16.67  & 4.212 & 10.00 \\
                   & Role + HR + ToT & 17.558 & 40.00  & 4.350 & 3.33 \\
\bottomrule
\end{tabular}
\end{table}

Table~\ref{tab:calibration_step2_robustness} shows that Tree-of-Thought (ToT) exhibits consistently higher or equal alignment with the expert reference decision across scenarios and explanation-length regimes, with the only exception occurring in Scenario~3 under extreme explanation compression. Importantly, in the most safety-critical condition (Scenario~2, oncoming traffic), ToT remains strongly aligned with expert judgment under unconstrained reasoning (93.33\% overtaking), whereas CoT collapses to 30\%.
This indicates that ToT preserves expert-aligned trade-off reasoning under normative conflict, while CoT becomes unstable when legality and safety constraints compete.

Because CARE-Drive evaluates reason-responsiveness rather than mere action selection, we additionally verify that both CoT and ToT satisfy a minimum normative validity condition: explicit acknowledgment that crossing a double solid line is illegal.

A post-hoc check of all 180 CoT and 180 ToT outputs confirms that all decision explanation $E_{VLM}^{(+R)}$ acknowledge the legal prohibition.
Thus, both strategies satisfy the normative-awareness requirement, and the final selection is based solely on decision robustness and expert alignment.

Based on superior robustness in safety-critical contexts and higher stability across explanation lengths, we select
\[
(M^*,T^*) = (\texttt{gpt-4.1},\text{Tree-of-Thought})
\]
as the calibrated prompt configuration for Stage~2.

\subsection{Stage 2: Contextual reasons evaluation (sensitivity)}

\subsubsection{Full-factorial results summary}

All results in this section use the calibrated configuration $(M^*,T^*)$ identified in Stage~1. The visual scene $V$ is fixed to Scenario~2 (oncoming vehicle), which represents the most safety-critical and normatively constrained overtaking condition. Consequently, all variation in this stage is introduced through the observable context $O$ and the explanation-length parameter $L$.

As defined in Table~\ref{tab:variables}, the observable context varies over $O \in \{TTC_o, B, U, F\}$, where $TTC_o$ denotes the time-to-collision with the oncoming vehicle, $B$ indicates the presence of a vehicle behind the AV, $U$ denotes passenger urgency, and $F$ represents the following time behind the cyclist. Explanation length is varied between $L \in \{\text{Few-Sentences},\ \text{No-Limit}\}$. The outcome variable is defined as
\[
Y = \mathbbm{1}\!\left[
D^{(+R)}_{VLM_{\texttt{gpt-4.1}}}(V_{\text{scenario-2}},O,L) = \text{Overtake}
\right],
\]
where $\mathbbm{1}[\cdot]$ denotes the indicator function. Thus, $Y=1$ corresponds to an overtaking decision and $Y=0$ corresponds to staying behind the cyclist.

For each unique combination of $(TTC_o, B, U, F, L)$, the $VLM_{\texttt{gpt-4.1}}$ was evaluated over 30 runs. The empirical overtake probability (i.e. overtaking rate) $P(Y=1)$ was computed as the proportion of runs in which $Y=1$. Figure~\ref{fig:overtaking_rate_plot} visualizes the overtaking rate as a function of $TTC_o$ for two explanation-length settings ($L=\text{Few-Sentences}$ and $L=\text{No-Limit}$) and three following-time conditions ($F \in \{12,18,24\}$). Each panel contrasts social-context conditions with and without a vehicle behind, as well as passenger urgency.

Across all conditions, increasing $TTC_o$ is generally associated with a higher overtaking probability, indicating sensitivity to available safety margins. However, this relationship depends on following time. For short following durations (F=12), overtaking rates generally increase with $TTC_o$, though minor non-monotonicities appear (e.g., under passenger urgency, rates at $TTC_o=8.5s$ slightly decrease compared to $TTC_o=6.8s$). For intermediate and long following durations ($F=18s$ and $F=24s$), the increase saturates and, in some cases, declines beyond mid-range $TTC_o$ values.

The presence of a vehicle behind the AV is consistently associated with higher overtaking probabilities across nearly all $TTC_o$ values. This effect is visible in Figure~\ref{fig:overtaking_rate_plot} as the systematic elevation of curves corresponding to $B=1$ relative to $B=0$. Contrary to our intuition, the addition of information about passenger urgency ($U=1$) reduces the likelihood of overtaking, with urgency-conditioned curves consistently lying below their neutral counterparts.

Interestingly, the length of explanation $L$ appears to have a dominant influence where when $L=\text{Few-Sentences}$ the overtaking rate of all combinations is almost 0, and there is an increase in the overtaking rate along with the increase in the following time, but it is still far compared to $L=\text{No-Limit}$, where overtaking rates vary substantially ($0-100\%$) across conditions, with near-zero rates only at very low $TTC_o$ values (1.7s).

Taken together, these patterns indicate that the calibrated VLM does not merely generate static decisions but systematically adjusts its behaviour in response to changes in observable context that are normatively relevant for human driving. Within the CARE-Drive framework, this constitutes evidence of context-sensitive reason-responsiveness, where injected human reasons interact with situational variables to shape the model’s decision preferences rather than serving as post-hoc rationalisations. Formal statistical analysis of these effects is provided in the subsequent subsection.

\subsubsection{Binary logit model results}

To quantify how observable context variables influence the calibrated VLM’s overtaking decisions, we estimated a binary logistic regression model using the full-factorial dataset described in the previous subsection. The dependent variable is defined as $Y \in \{0,1\}$, where $Y=1$ denotes an overtaking decision and $Y=0$ denotes staying behind the cyclist. The independent variables correspond to the observable context components $O=\{TTC_o, B, U, F\}$ and the explanation-length regime $L$.

The estimated coefficients and associated statistics are reported in Table~\ref{tab:resultsbinarylogit}. The intercept $\beta_0=-1.953$ indicates that, under baseline conditions (i.e., minimal $TTC_o$, no vehicle behind, no passenger urgency, shortest following time, and constrained explanation length), the model exhibits a low intrinsic probability of overtaking. This baseline probability corresponds to approximately 12.4\%, as shown in Table~\ref{tab:resultsprobability}.

\begin{figure}
    \centering
    \includegraphics[width=1\linewidth]{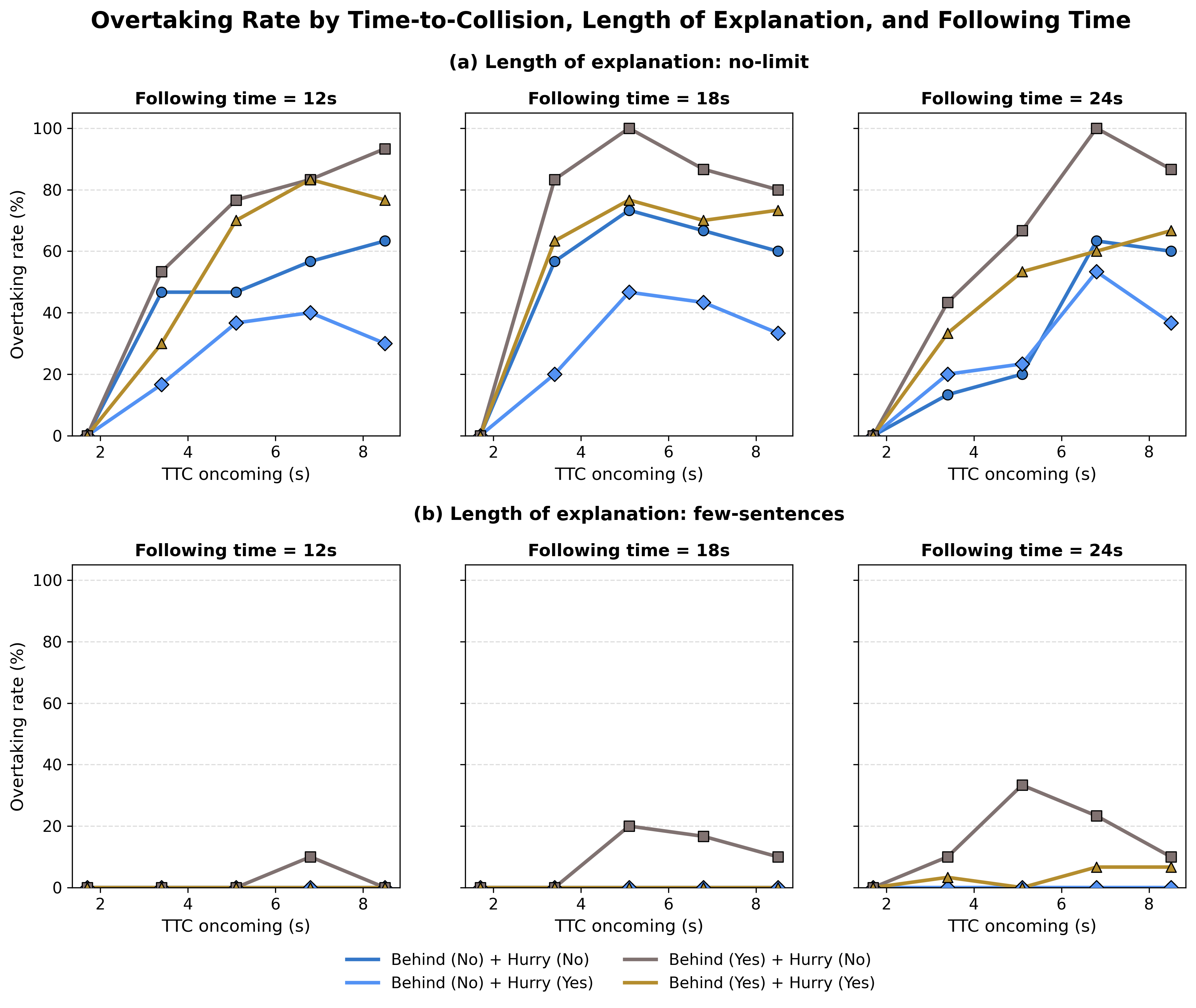}
    \caption{Full-factorial empirical overtaking probability $P(Y=1)$ of the calibrated CARE-Drive configuration, where $(M^*,T^*)=(\texttt{gpt-4.1},\text{Tree-of-Thought})$, under systematic variation of observable context and explanation length. Each subplot shows the empirical overtaking probability $P(Y=1)$ as a function of time-to-collision $TTC_o$. Columns correspond to following time $F\in\{12,18,24\}\,$s, and rows correspond to the explanation-length regime (top row: $L=\text{No-Limit}$; bottom row: $L=\text{Few-Sentences}$). Within each subplot, curves represent combinations of vehicle-behind indicator $B\in\{0,1\}$ and passenger urgency $U\in\{0,1\}$. Each point represents the proportion of overtaking decisions over 30 stochastic runs per condition.
}

    \label{fig:overtaking_rate_plot}
\end{figure}

\begin{table*}[ht]
\centering
\caption{Binary logit regression results showing the effect of observable context and explanation length on overtaking decisions.}
\label{tab:resultsbinarylogit}
\begin{tabular*}{\linewidth}{@{\extracolsep{\fill}}lccccc}
\hline
 & \textbf{Coefficient ($\beta$)} & \textbf{Significance} & \textbf{Odds ratio} & \textbf{95\% CI ($\beta$)} & \textbf{p-value} \\
\hline
Intercept & $-1.953$ & *** & 0.142 & [-2.238, -1.669] & $<0.001$ \\
Time-to-collision ($TTC_o$) & $3.015$ & *** & 20.389 & [2.687, 3.343] & $<0.001$ \\
Vehicle behind ($B$) & $1.330$ & *** & 3.781 & [1.116, 1.545] & $<0.001$ \\
Passenger urgency ($U$) & $-0.872$ & *** & 0.418 & [-1.081, -0.663] & $<0.001$ \\
Following time ($F$) & $-0.049$ &  & 0.952 & [-0.299, 0.201] & 0.702 \\
Explanation length ($L$) & $-4.184$ & *** & 0.015 & [-4.529, -3.840] & $<0.001$ \\
\hline
\multicolumn{6}{l}{\footnotesize Significance levels: *** $p<0.001$, ** $p<0.01$, * $p<0.05$.}
\end{tabular*}
\end{table*}

Among all predictors, time-to-collision with the oncoming vehicle ($TTC_o$) has the strongest positive effect on overtaking likelihood ($\beta_1=3.015$, $p<0.001$). This indicates that larger safety margins substantially increase the probability of overtaking. The corresponding odds ratio of 20.389 implies that increasing $TTC_o$ from its minimum to maximum normalized value increases the odds of overtaking by more than a factor of twenty. This finding suggests that larger safety gaps drives overtaking decisions. The presence of a vehicle behind the AV also significantly increases overtaking likelihood ($\beta_2=1.330$, $p<0.001$), with an odds ratio of 3.781. This suggests that the model responds to rear-vehicle pressure.

\begin{table}[h]

\centering

\caption{Predicted overtaking probability under baseline conditions and individual predictor changes}

\label{tab:resultsprobability}

\begin{tabular}{lcc}

\hline

\textbf{Predictor condition} & \textbf{Predicted probability} & \textbf{Change from baseline} \\

\hline

Baseline ($TTC_o = 0, B = 0, U = 0, F = 0, L = 0$) & 12.4\% & --- \\

Time-to-collision increased ($TTC_o = 1\text{, normalized}$) & 74.3\% & $\textcolor{teal}{+61.9\%}$ \\

Vehicle behind present ($B = 1$) & 34.9\% & $\textcolor{teal}{+22.5\%}$ \\

Passenger urgency present ($U = 1$) & 5.6\% & $\textcolor{purple}{-6.8\%}$ \\

Following time increased ($F = 1\text{, normalized}$) & 11.9\% & $\textcolor{purple}{-0.5\%}$ \\

Explanation present ($L = 1$) & 0.2\% & $\textcolor{purple}{-12.2\%}$ \\

\hline

\end{tabular}

\end{table}

In contrast, passenger urgency ($U$) significantly reduces overtaking likelihood ($\beta_3=-0.872$, $p<0.001$), corresponding to an odds ratio of 0.418. This result indicates that when urgency information is present, the model becomes more conservative rather than more aggressive. Following time ($F$) exhibits a small negative coefficient ($\beta_4=-0.049$), but this effect is not statistically significant ($p=0.702$). This suggests that, after controlling for other variables such as safety margin and social pressure, following time alone does not independently influence the overtaking decision.

Explanation length ($L$) has a large negative coefficient ($\beta_5=-4.184$, $p<0.001$), indicating that constrained explanation regimes substantially reduce overtaking likelihood. The corresponding odds ratio of 0.015 shows that limiting reasoning bandwidth strongly suppresses overtaking decisions. The resulting fitted logistic regression model is given by:
\begin{equation}
\label{eq:logit_filledin}
\begin{aligned}
\text{logit}(p)
=
-1.953
+ 3.015\,TTC_o
+ 1.330\,B
- 0.872\,U
- 0.049\,F
- 4.184\,L
\end{aligned}
\end{equation}

where $p=P(Y=1)$ denotes the probability of overtaking.

To facilitate interpretation, Table ~\ref{tab:resultsprobability} reports the predicted overtaking probabilities under baseline conditions and under changes in predictors. Increasing $TTC_o$ from $1.7s$ to $8.5s$ results in the largest absolute probability increase, from 12.4\% to 74.3\%, confirming that the safety margin is the dominant factor influencing decisions. The presence of a rear vehicle increases the overtaking probability to 34.9\%, reflecting the model's sensitivity to the social driving context. Passenger urgency reduces the overtaking probability to 5.6\%, while the explanatory length constraint suppresses overtaking even more, reducing the probability to about 0.2\%. These results quantitatively complement the qualitative trends observed in the empirical probability curves in Figure ~\ref{fig:overtaking_rate_plot}.

Overall, these results indicate that the calibrated VLM configuration exhibits systematic and statistically significant sensitivity to normatively relevant contextual variables. Specifically, safety margin (time-to-collision) and social context (presence of a vehicle behind) significantly increase the likelihood of overtaking, while explanation-length constraints substantially decrease the likelihood of overtaking. However, following time does not significantly affect the likelihood of overtaking, and information indicating that the passenger is in a hurry unexpectedly reduces the likelihood of overtaking. These findings suggest that VLM decision-making behavior is not static, but instead responds systematically to observable changes in context.

\subsubsection{CARLA validation}

To assess whether the calibrated CARE-Drive configuration can be integrated into a dynamic driving system, we implemented the optimal prompt configuration $(M^*,T^*)=(\texttt{gpt-4.1},\text{Tree-of-Thought})$ in the CARLA simulator.

In this setup, a simulated automated vehicle approaches a cyclist in Scenario 1 (None) and Scenario 2 (Oncoming). While awaiting the VLM decision, the AV is programmed to maintain the cyclist's speed to ensure safe following behavior. The scene representation follows the observable context value $O$ used in the CARE-Drive evaluation in Step 1. To ensure the decision-making process is not too lengthy, we use an explanation length of $L=\text{Few-Sentences}$.

To reduce stochastic variability, each decision query is executed twice. If both executions result in the same decision, the maneuver is executed. In case of disagreement, a third query resolves the decision by majority vote. In all tested scenarios, the model produced consistent decisions without requiring deadlock resolution. To ensure that the decision-making process could be executed without undue delay,

The CARLA results qualitatively matched the CARE-Drive findings. When there were no oncoming vehicles (Scenario 1), the model consistently chose to overtake. When there were oncoming vehicles (Scenario 2), the model consistently chose to stay behind. These results confirm that the calibrated CARE-Drive configuration produces stable decisions that can be directly translated into executable vehicle behavior in a closed simulation environment.

This experiment validates the feasibility of CARE-Drive's decision behavior, demonstrating that the framework's calibrated decision output is not only statistically consistent but also operationally executable in dynamic driving scenarios. A video demonstration of the simulation is available online\footnote{\url{https://elsefientulleners.wixsite.com/bep9}}.

\section{Discussion}
\subsection{Key findings}
This study investigates whether a vision-language model (VLM) embedded with explicit human-centered reasons in its prompt, when evaluated using the CARE-Drive framework, demonstrates responsiveness to normative reasons in ethically ambiguous driving scenarios. Our findings indicate that VLM decisions are responsive to normative reasons to a meaningful degree, although this responsiveness is not uniform across all tested reasons.

To operationalize normative reasons, we identified observable contextual variables $O$ grounded in human factors and driving behavior research. Specifically, Time-to-Collision ($TTC_o$), presence of a vehicle behind $B$, passenger urgency $U$, and following time $F$ were selected as contextual indicators corresponding to safety, efficiency, and comfort considerations. These variables, commonly studied in human driving research, serve as observable proxies for normative considerations. Importantly, this operationalization does not assume that VLM decision-making is equivalent to human cognition; rather, human factors findings are used to interpret the normative meaning of contextual variables within driving scenarios.

We found that the VLM demonstrates responsiveness to safety-related reasons. In particular, higher $TTC_o$ values were associated with increased overtaking probability, consistent with evidence that time-to-collision reflects the safety margins drivers maintain during overtaking \citep{kovaceva2019drivers}. Similarly, larger gaps to oncoming vehicles, corresponding to higher TTC under comparable speeds, are more likely to be accepted for overtaking \citep{sevenster2023response}. In addition, the presence of a following vehicle $(B=1)$ increased overtaking probability, aligning with findings that closely following vehicles create social pressure on drivers ahead to accelerate or change lanes, thereby motivating overtaking to maintain traffic flow \citep{bumrungsup2022analysis}. This suggests that the model adjusts its decision behavior in ways consistent with safety-related normative considerations.

However, responsiveness was not consistent across all reason categories. Contrary to expectations, introducing passenger urgency, an indicator of efficiency-related reasons, did not increase overtaking probability. Instead, the model exhibited lower overtaking probability compared to baseline conditions. This contrasts with prior findings showing that time pressure increases drivers’ willingness to engage in more assertive or risk-accepting behaviors to achieve efficiency goals \citep{yao2026driving}. Similarly, increasing following duration was intended to operationalize accumulated delay and driver impatience, which have been shown to increase overtaking intentions in human drivers due to frustration and efficiency-related motivational factors \citep{kinnear2015experimental}. However, the VLM did not exhibit increased overtaking propensity under this condition. This may reflect a more conservative decision strategy, as prior work has shown that prolonged delay does not universally increase overtaking; while younger drivers tend to overtake more readily under delay, older drivers often exhibit more cautious behavior and are less likely to accept overtaking opportunities under the same conditions \citep{llorca2013influence}. This contrasts with evidence that human overtaking behavior also reflects sensitivity to social and comfort-related considerations of other road users, such as maintaining lateral comfort distances when overtaking cyclists \citep{farah2019modelling}. The absence of responsiveness to delay-based efficiency cues suggests that the model may prioritize safety-related considerations over efficiency-related motivations when these factors do not directly alter safety constraints. Together, these results indicate that while the VLM responds to certain normative reasons, its responsiveness remains selective and uneven.

In addition, explicitly including a structured list of normative reasons in the prompt, together with reasoning strategies such as Chain-of-Thought and Tree-of-Thought prompting, increased empirical overtaking probability in ethically ambiguous situations. In many such cases, overtaking was the action preferred by expert evaluators. This suggests that making normative reasons explicit can improve alignment between VLM decision recommendations and expert-preferred driving behavior. 

Taken together, these findings demonstrate that VLM decision outputs are sensitive to normative reasoning cues and contextual conditions, although this sensitivity varies across different types of reasons.

\subsection{Relationship to prior work on VLM reasoning}
These findings address an important limitation in prior VLM-based automated driving research, which has primarily focused on improving reasoning generation rather than evaluating reason responsiveness. Earlier approaches, including Reason2Drive \citep{nie2024reason2drive}, Alpamayo-R1\citep{wang2025alpamayo}, and related methods, were developed to move beyond direct question–answering formulations by encouraging models to generate structured reasoning alongside their decisions. These methods demonstrated that VLMs can produce plausible and coherent explanations. However, their primary objective was to improve reasoning quality and interpretability, rather than to establish whether normative reasons systematically influence decision outcomes. In particular, because reasoning was generated as part of the model’s output without controlled variation of normative inputs, it remained unclear whether the provided reasons were functionally connected to the decision or merely post-hoc justifications.

This distinction is particularly important from the perspective of meaningful human control \citep{mecacci2020meaningful}. Under meaningful human control, autonomous system behavior should remain responsive to the human-relevant reasons that justify its actions. Generating plausible explanations alone is insufficient if those explanations do not actually guide decision outcomes. If a system produces explanations that appear normatively appropriate but are not functionally connected to its decisions, this can create misplaced trust, as human operators may assume that the system’s behavior is guided by appropriate normative considerations when it is not.

This concern is especially relevant for models trained using human-annotated driving data derived from observed actions in videos. In such training paradigms, annotations typically describe or rationalize observed behavior after the fact, rather than explicitly linking decisions to the normative reasons that justify them. As a result, models trained in this manner may learn to reproduce plausible explanatory patterns without establishing a functional relationship between reasons and decisions. This can lead to explanation–decision decoupling, where explanations reflect human-like narratives while decisions remain driven by other latent factors learned during training.

CARE-Drive complements this line of work by introducing a structured evaluation framework designed specifically to probe reason responsiveness. Rather than modifying the model architecture or training paradigm, CARE-Drive introduces explicit normative reasons into the prompt and systematically examines how decision outputs change under controlled contextual conditions. This enables empirical assessment of whether VLM decisions exhibit systematic relationships with normative reasoning inputs. While CARE-Drive does not redesign the model from first principles to enforce reason-driven decision making, it provides an intermediate step toward evaluating reason responsiveness using existing models. This distinction allows CARE-Drive to assess the behavioral relationship between reasons and decisions, without assuming that reasoning generation alone implies reason-driven decision making.

Importantly, CARE-Drive does not claim causal relationships between reasons and decisions. Instead, it provides empirical evidence of systematic associations between explicit normative reasoning inputs and observable changes in VLM decision behavior.

\subsection{Limitations and Future Work}

Several limitations of this study should be acknowledged.

First, the operationalization of normative reasons relies on proxy contextual variables rather than direct access to the model’s internal reasoning processes. CARE-Drive evaluates observable relationships between normative reason inputs and decision outputs, but does not establish whether the model internally represents or causally uses those reasons. As a result, the framework provides behavioral evidence of reason responsiveness rather than direct insight into the model’s internal reasoning mechanisms. Future research could extend this work by combining CARE-Drive with training paradigms or architectures explicitly designed to link decisions to structured normative reasoning, enabling closer examination of how reasons are represented and used within the model.

Second, the evaluation relies on prompt-based reason specification and instruction calibration. While CARE-Drive systematically controls prompt structure and model configuration, the observed reason responsiveness may depend on how normative reasons are expressed in the prompt. Future work could investigate alternative formulations of reason representation, including structured symbolic inputs or multimodal reasoning representations, to assess robustness across different prompting strategies.

Third, the number of repeated simulations per condition is limited. Increasing the number of repetitions would improve statistical robustness, reduce variance in empirical probability estimates, and enable more precise characterization of reason responsiveness across contextual conditions.

Fourth, the reasoning strategy used in the evaluation was selected through the prompt calibration process, which prioritized alignment between model decisions and the expert reference outcome. In this study, the tree-of-thought (ToT) strategy demonstrated higher alignment with the expert-recommended overtaking decision compared to chain-of-thought (CoT). However, this selection criterion is based on behavioral alignment rather than direct evaluation of reasoning fidelity or causal use of normative reasons. As a result, the observed reason responsiveness may partially depend on the reasoning strategy selection process. Future work could evaluate multiple reasoning strategies in parallel or develop calibration criteria that explicitly assess the quality and causal role of normative reasoning, rather than relying solely on decision alignment.

Finally, the experimental evaluation focused on overtaking cyclists, a scenario chosen for its well-documented normative trade-offs between safety, efficiency, and social considerations. However, automated driving involves a wide range of ethically ambiguous situations. Future work should apply CARE-Drive to additional driving contexts, such as merging, yielding, pedestrian interactions, and multi-agent traffic environments, to assess the generality of the observed reason–decision relationships.

\section{Conclusion}
This study introduced CARE-Drive, a structured framework for evaluating reason-responsive decision behavior in vision–language models applied to automated driving. Our findings demonstrate that VLM decision outputs exhibit measurable responsiveness to normative reasons and contextual conditions, particularly for safety-related factors. However, this responsiveness is not uniform across different categories of reasons, indicating that the model selectively responds to certain normative considerations while remaining less sensitive to others.

These results provide new empirical evidence that explicitly presenting normative reasons can influence VLM decision recommendations in ethically ambiguous driving scenarios. Importantly, CARE-Drive enables structured assessment of this relationship by systematically varying contextual variables that operationalize normative considerations grounded in human driving research. This allows behavioral evaluation of reason responsiveness without requiring direct access to the model’s internal reasoning mechanisms.

By enabling structured analysis of the relationship between normative reasons, contextual conditions, and decision outputs, CARE-Drive provides a practical method for diagnosing whether autonomous decision systems behave in ways consistent with reason-responsive principles. This contributes toward the broader goal of ensuring that automated driving systems remain aligned with human-relevant normative considerations. More broadly, CARE-Drive establishes a foundation for future research on reason-aware evaluation of AI decision-making systems in safety-critical domains.

\section*{Code Availability}

The CARE-Drive framework implementation, including prompt calibration, evaluation scripts, and analysis code, is publicly available at: \url{https://github.com/lucassuryana/CARE-Drive}. The repository contains all scripts required to reproduce the prompt calibration procedure, contextual sensitivity evaluation, and statistical analysis presented in this study.

\section*{Acknowledgement}
The authors acknowledge the use of ChatGPT as a language assistance tool to improve clarity, grammar, and overall readability of portions of the manuscript. All scientific content, analysis, and conclusions are solely the responsibility of the authors.


\bibliographystyle{model5-names}
\bibliography{cas-refs}





\end{document}